\def\year{2021}\relax
\documentclass[letterpaper]{article}
\usepackage{aaai21}
\usepackage{times}
\usepackage{helvet}
\usepackage{courier}
\usepackage[hyphens]{url}
\usepackage{graphicx} 
\usepackage{graphicx}
\usepackage{natbib}
\usepackage{caption}
\usepackage{bm}
\usepackage{bbm}
\usepackage{amsmath,amsfonts,amsthm,amssymb}
\usepackage{xcolor}

\usepackage[caption=false,font=footnotesize,hangindent=10pt]{subfig}
\graphicspath{{./}{pics/}}
\usepackage{tikz}
\usetikzlibrary{calc,positioning,fit}
\usepackage{appendix} 
\newcommand{\transpose}{^{\ensuremath{\mathsf{T}}}}

\newcommand{\bullethdr}[1]{\smallskip\noindent\textbullet\,\textbf{#1}}

\newcommand{\Real}{\mathbb{R}}
\newcommand{\be}{\mathbf{e}}
\newcommand{\bb}{\mathbf{b}}

\newcommand{\bu}{\mathbf{u}}
\newcommand{\bbm}{\mathbf{m}}
\newcommand{\bq}{\mathbf{q}}

\newcommand{\balpha}{\bm{\alpha}}
\newcommand{\bbeta}{\bm{\beta}}

\newcommand*\samethanks[1][\value{footnote}]{\footnotemark[#1]}

\begin{document}

\title{Learning to Check Contract Inconsistencies}
\author {
  Shuo Zhang\textsuperscript{\rm 1},
  Junzhou Zhao\textsuperscript{\rm 1}\thanks{ Corresponding Author },
  Pinghui Wang\textsuperscript{\rm 1}\samethanks[1], 
  Nuo Xu\textsuperscript{\rm 1},
  \\
  Yang Yang\textsuperscript{\rm 1},
  Yiting Liu\textsuperscript{\rm 1},
  Yi Huang\textsuperscript{\rm 2},
  Junlan Feng\textsuperscript{\rm 2} 
  \\
}
\affiliations {
    % Affiliations
    \textsuperscript{\rm 1} MOE KLINNS Lab, Xi'an Jiaotong University, Xi'an 710049, P. R. China \\
    \textsuperscript{\rm 2} JIUTIAN Team, China Mobile Research\\
    \{zs412082986, l530111398\}@stu.xjtu.edu.cn, \{junzhouzhao, yang.yang95868\}@gmail.com\\phwang@mail.xjtu.edu.cn, nxu@sei.xjtu.edu.cn, \{huangyi, fengjunlan\}@chinamobile.com\\
}
\maketitle

\begin{abstract}
Contract consistency is important in ensuring the legal validity of the contract.
In many scenarios, a contract is written by filling the blanks in a precompiled
form.
Due to carelessness, two blanks that should be filled with the same (or different) content may
be incorrectly filled with different (or same) content.
This will result in the issue of {\em contract inconsistencies}, which may
severely impair the legal validity of the contract.
Traditional methods to address this issue mainly rely on manual contract review,
which is labor-intensive and costly.
In this work, we formulate a novel {\em Contract Inconsistency Checking} (CIC)
problem, and design an end-to-end framework, called {\em Pair-wise Blank
  Resolution} (PBR), to solve the CIC problem with high accuracy.
Our PBR model contains a novel \texttt{BlankCoder}
to address the challenge of modeling meaningless blanks.
\texttt{BlankCoder} adopts a two-stage attention mechanism that adequately associates a meaningless blank with its relevant descriptions while avoiding the incorporation of irrelevant context words.
Experiments conducted on real-world datasets show the promising performance of our
method with a balanced accuracy of $94.05\%$ and an F1 score of $90.90\%$ in the CIC
problem.
\end{abstract}

\section{Introduction}

A contract is a legally binding agreement that recognizes and governs the rights
and duties of the parties to the agreement.
Correctly composing contracts is crucial to ensure its legal validity.
In many real-world scenarios, a standard contract is prepared by {\em filling
  blanks} in a precompiled form.
Due to carelessness, two blanks that should be filled with the same (or different) content may be incorrectly filled with different (or same) content.
This will result in \emph{contract inconsistencies}, which may severely impair the
legal validity of the contract.

Contract review is widely used by companies to check contract inconsistencies.
However, contract review is labor-intensive and costly.
Big companies have to hire tens of thousands of lawyers to conduct contract
review, and it is estimated that Fortune Global $2000$ and Fortune $1000$
companies spend about $\$35$ billion annually to review and negotiate
contracts~\cite{Strom2019}.
Therefore, it is desired to design methods to automate the contract review
process.

In this work, we study the {\em Contract Inconsistency Checking} (CIC) problem,
and propose an end-to-end model to solve the CIC problem with high accuracy.
As far as we know, the CIC problem has not yet been studied in the AI community
and no effective solution exists.

\begin{figure}[t]
	
	\includegraphics[width=.4\textwidth]{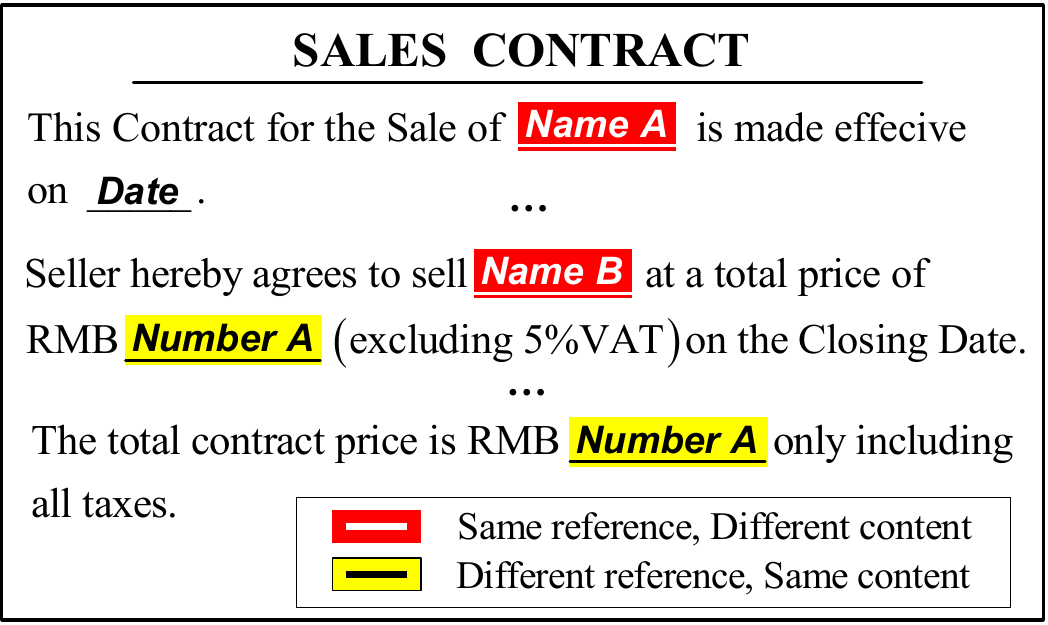}
	\centering
	\caption{Examples of contract inconsistencies.
		The two blanks with red background refer to the same item for sale but
		incorrectly filled with different content.
		The two blanks with yellow background refer to two different prices (one with
		tax and the other without tax) but incorrectly filled with the same price.}
	\label{fig:examplecase}
\end{figure}

We consider contract inconsistencies caused by incorrectly filling blanks in a
precompiled form.
For example, in Figure~\ref{fig:examplecase}, the two blanks with red background
actually refer to the same item for sale, but incorrectly filled with different
content.
The two blanks with yellow background actually refer to two different prices
(i.e., one with tax and the other without tax), but incorrectly filled with the
same price.
Our goal is to find methods that can automatically detect such contract
inconsistencies with high accuracy.

A straightforward method to solve the CIC problem is to build a set of rules for
each contract, and these rules regularize the allowed content for each blank.
Rule-based methods have been used for other contract-related tasks such as
checking obligation violations in contracts~\cite{Governatori2005}, verifying
electronic contracts~\cite{Abdelsadiq2011}, and contract
formalization~\cite{Joanni2018}.
However, the rules have to be built manually by experts,
and they do not scale well to diverse types of contracts.

Coreference resolution (CR) methods~\cite{Sukthanker2020} aim to identify words or
phrases that refer to the same real-world entity in a document.
At first glance, CR methods may solve the CIC problem if we view blanks as words
or phrases, and hence CR methods can be used to answer whether two blanks refer to
the same concept.
However, blanks are indeed not words or phrases which can be encoded using well
studied word2vec methods, while blanks are meaningless empties and have no
semantic meaning at all.
Existing CR methods fail in modeling blanks in our case.

In this work, we propose a {\em Pair-wise Blank Resolution} (PBR) framework to solve
the CIC problem with high accuracy.
We formulate the CIC problem as a pair-wise binary classification problem.
For a pair of blanks in the contract document, our PBR model adopts the Siamese
architecture~\cite{Bromley1994,Reimers2019} to encode each blank separately
through the same blank encoder and then predict whether they should be filled with
the same content or not.
Blank modeling is challenging since it is hard to perform a semantic comparison to associate a meaningless blank to its relevant descriptions.
In our PBR framework, we propose a novel \texttt{BlankCoder} that extends the
Transformer~\cite{Vaswani2017} encoder architecture to address the above issue.
\texttt{BlankCoder} adopts a two-stage modeling strategy where a blank is first initialized with the more related local context words and then
updated by recurrently incorporating relevant information from the global context words.
In this way, related descriptions of the blank would be fully introduced and the irrelevant ones would be ignored, yielding an informative blank representation.

To evaluate our method on the CIC task, we build a large-scale Chinese contract dataset of $281$ contract documents
including $299,621$ training samples (blank pairs).
The English contract dataset for element extraction released by Chalkidis et
al.~\shortcite{Chalkidis2017} is also used, and we view each element as
a filled blank.
The experimental results show that our method significantly and consistently
outperforms all baseline methods with a promising performance of the balanced accuracy of $94.05\%$ and F1 of
$90.90\%$.

Our contributions are summarized as follows:

(1) We formulate the {\em Contract Inconsistency Checking} (CIC) problem.
As far as we know, this problem has not yet been studied in the AI community.

(2) We propose a novel {\em Pair-wise Blank Resolution} (PBR) framework to address the CIC problem.
In PBR, we propose a {\em \texttt{BlankCoder}} that extends the Transformer encoder architecture to efficiently model meaningless blanks.

(3) We collected and labeled 
a large-scale Chinese contract corpus for CIC.
The experimental results show the promising performance of our PBR method.

\section{Related Work}

Our work is mainly related to three lines of recent researches: {\em automatic contract analysis}, {\em coreference resolution}, and {\em blank modeling}.

Existing automatic contract analysis methods mainly assist legal professionals by information extraction.
Early ML-based attempts performed sentence-level classification on clause patterns~\cite{indukuri2010mining} and service exceptions~\cite{gao2011mining}.
Recent DL-based methods focus on fine-grained intra-document classification of
contract elements~\cite{chalkidis2019neural}, obligations and prohibitions~\cite{chalkidis2018obligation}, and insurance policies~\cite{sun2019toi}, and cross-document search of relevant clauses~\cite{guodeep}.
Previous works do not perform a further comparison on the retrieved related sentences, leaving the detailed checking process to lawyers.
% are restricted to finding related sentences of the one under review, leaving further reading comprehension to legal professionals. 
In this paper, we automate the CIC process in a fully data-driven and end-to-end manner that significantly speeds up the manual review process.

Coreference resolution (CR) aims to identify the words or phrases (mentions) that refer to the same real-world entity.
Existing methods can be classified into three broad
categories of mention-pair, entity-mention, and ranking models.
Mention-pair models~\cite{wiseman2015learning} predict the coreference label for every two mentions.
Entity-mention models~\cite{clark2016improving} directly model an entity by clustering mentions.
Ranking models~\cite{lee2017end} were further introduced to model the degree of coreference.
CIC can be view as a modified CR task that aims to identify the blanks that refer to the same content.
However, CIC is much more challenging since the blanks are meaningless empties that can not be addressed with CR methods.

Blank modeling has been investigated in Text Infilling and Zero Pronoun Resolution (ZPR).
In text infilling, a blank is usually treated as an out-of-vocabulary token and modeled with sequence models such as BiLSTM~\cite{fedus2018maskgan} and Transformer~\cite{zhu2019text}.
However, these methods encode a blank with all its context words that contain irrelevant noise descriptions. 
Similar to CR, ZPR aims to identify words that co-refer with a gap (an omitted pronoun).
To encode the gap, Yin et al.~\shortcite{Yin2016} designed a CenteredLSTM that focuses on the more related local words.
In their later work~\cite{yin2018zero}, self-attention was further utilized to enhance the performance.
Recently, Aloraini et al.~\shortcite{Aloraini2020} adopted BERT~\cite{Devlin2019} to encode the gap with its nearest two words.
Though able to avoid incorporating irrelevant descriptions, ZPR methods would negligent faraway relevant descriptions.
% yield inadequate blank representations due to the negligence of 

\section{Contract Inconsistency Checking Problem}

We formulate the {\em Contract Inconsistency Checking} (CIC) problem as a
pair-wise binary classification problem, i.e., given a pair of blanks occurred in
the contract document, we want to predict whether they should be filled with the
same content or not.

Formally, let $B$ denote a set of blanks in the contract document.
Each blank $b\in B$ is included in a {\em surrounding sentence}
$s=\{w_1,w_2,\ldots,b,\ldots\}$ consisting of $|s|-1$ words and one
blank\footnote{If a sentence in the contract document contains multiple blanks, we
  construct the surrounding sentence $s$ by deleting the other blanks and only
  keep the blank of interest and all the words in the original sentence.}.
Given two blanks $b_i, b_j\in B$, the CIC problem aims at predicting their {\em
  consistency relation} $r_{ij}\in\{0,1\}$, with $r_{ij}=1$ meaning that the two
blanks $b_i$ and $b_j$ should be filled with the same content; and $r_{ij}=0$
otherwise.

\section{Pair-wise Blank Resolution Framework}

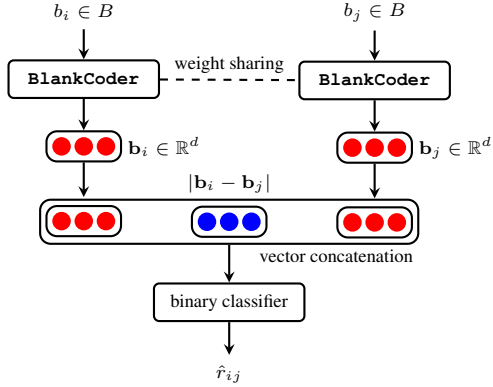
\begin{figure}[htp]
  \centering
  \begin{tikzpicture}[
  font=\scriptsize,
  txt/.style={align=center,inner sep=2pt},
  rec/.style={draw,thick,minimum width=2cm,minimum height=5mm,rounded corners=2pt,text depth=.25ex},
  enc_rec/.style={draw,thick,rounded corners=4pt,inner sep=2pt},
  arr/.style={-stealth, thick},
  cir/.style={circle,fill=red,minimum size=7pt,inner sep=0pt},
  ]
  \node (Bn) {$b_i\in B$};
  \node[right=2.8 of Bn] (Bm) {$b_j\in B$};

  \node[rec,below=.4 of Bn] (BCn) {\textbf{\texttt{BlankCoder}}};
  \node[rec,below=.4 of Bm] (BCm) {\textbf{\texttt{BlankCoder}}};

  % connections
  \draw[dashed,thick] (BCn) to node[txt,above] {weight sharing}  (BCm) ;
  \draw[arr] (Bn) -- (BCn);
  \draw[arr] (Bm) -- (BCm);

  % bn
  \node[cir,below=.5 of BCn] (cn1) {};
  \node[cir,left=1pt of cn1] (cn0) {};
  \node[cir,right=1pt of cn1] (cn2) {};
  \node[enc_rec,fit={(cn0) (cn2)}] (bn) {};

  \node[txt,right=0 of bn] {$\bb_i\in\Real^d$};
  \draw[arr] (BCn) -- (bn);

  % bm
  \node[cir,below=.5 of BCm] (cm1) {};
  \node[cir,left=1pt of cm1] (cm0) {};
  \node[cir,right=1pt of cm1] (cm2) {};
  \node[enc_rec,fit={(cm0) (cm2)}] (bm) {};

  \node[txt,right=0 of bm] {$\bb_j\in\Real^d$};
  \draw[arr] (BCm) -- (bm);

  % encodings
  %% bn
  \node[cir,below=.65 of bn] (cn1) {};
  \node[cir,left=1pt of cn1] (cn0) {};
  \node[cir,right=1pt of cn1] (cn2) {};
  \node[enc_rec,fit={(cn0) (cn2)}] (bn_left) {};
  \coordinate[above=3pt of bn_left] (A);

  \draw[arr] (bn) -- (A);

  %% bm
  \node[cir,below=.65 of bm] (cm1) {};
  \node[cir,left=1pt of cm1] (cm0) {};
  \node[cir,right=1pt of cm1] (cm2) {};
  \node[enc_rec,fit={(cm0) (cm2)}] (bm_right) {};
  \coordinate[above=3pt of bm_right] (B);

  \draw[arr] (bm) -- (B);

  %% |bn - bm|
  \node[cir,fill=blue] (c1) at ($(cn1)!0.5!(cm1)$) {};
  \node[cir,fill=blue,left=1pt of c1] (c0) {};
  \node[cir,fill=blue,right=1pt of c1] (c2) {};
  \node[enc_rec,fit={(c0) (c2)}] {};

  \node[enc_rec,fit={(bn_left) (bm_right)}] (b) {};
  \node[txt,above=0 of b] {$|\bb_i-\bb_j|$};
  \node[txt,below=2pt of bm_right.south west] {vector concatenation};

  % classfier
  \node[rec,below=.5 of b] (c) {binary classifier};
  \draw[arr] (b) -- (c);

  \draw[arr] (c) -- ++(0,-.7) node[below] {$\hat r_{ij}$};
\end{tikzpicture}

%%% Local Variables:
%%% mode: latex
%%% TeX-master: "main"
%%% End:
  \caption{The PBR framework.}
  \label{fig:PBR}
\end{figure}

In this section, we propose a {\em Pair-wise Blank Resolution} (PBR) framework to
solve the CIC problem.
The overall structure of our PBR framework is illustrated in Figure~\ref{fig:PBR}.

\subsection{Overview of the PBR Framework}

The PBR framework is inspired by {\em Coreference Resolution}
(CR)~\cite{Sukthanker2020}.
CR aims at identifying words or phrases that refer to the same real-world entity
in a document.
However, as we explained previously, blanks are different from words or phrases
because blanks have no semantic meaning at all.
The key challenge for CIC is to represent the meaningless blanks in a contract
document.

As illustrated in Figure~\ref{fig:PBR}, the proposed PBR framework adopts the
Siamese architecture with a novel \texttt{BlankCoder} model to encode each blank, and
predicts the consistency relation of the blank pair by a binary classifier.
The proposed \texttt{BlankCoder} is a Transformer variant for blank modeling that could easily generalize to other tasks such as text infilling.

\subsection{\texttt{BlankCoder}}

The Transformer architecture~\cite{Vaswani2017} has shown a powerful semantic
modeling capability on various natural language processing tasks including the
relevant task of text infilling~\cite{zhu2019text}.
We extend the Transformer encoder architecture, and propose \texttt{BlankCoder} to
address the challenge of modeling meaningless blanks.

In \texttt{BlankCoder}, in order to obtain a good blank embedding, each blank is
made to fully utilize its context information in the surrounding sentence using
{\em a two-stage modeling strategy}, i.e., the blank is first initialized based on
the local context information (e.g., words in a local region of the blank), and
then refined and updated based on the global context information (e.g., words
outside of the local region in the surrounding sentence).

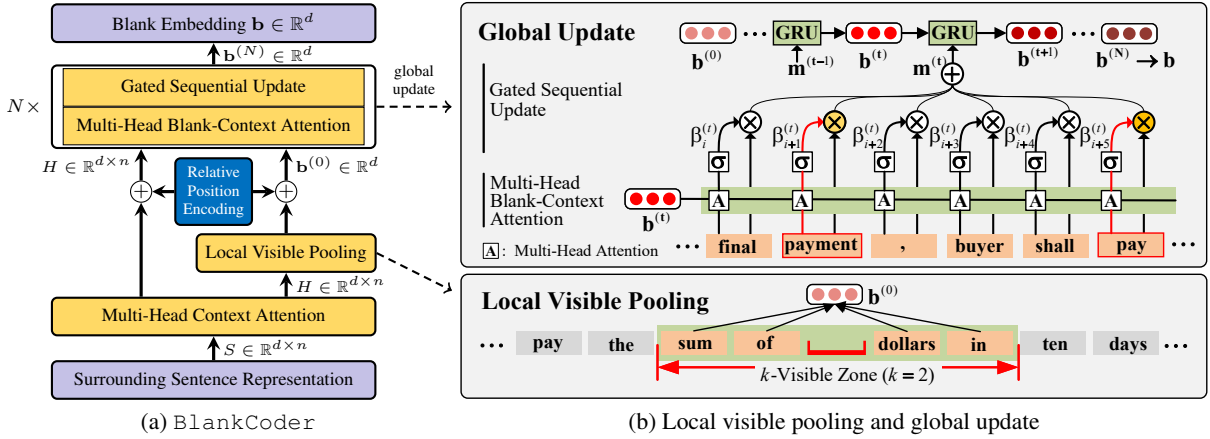
\begin{figure*}
  \centering
  \subfloat[\texttt{BlankCoder}\label{fig:blankcoder}]{\begin{tikzpicture}[
  font=\scriptsize,
  txt/.style={align=center,inner sep=3pt,font=\tiny},
  myrec/.style={draw,thick,minimum width=4.3cm,minimum height=2mm,rounded corners=2pt,text depth=.25ex},
  rec/.style={draw,thick,minimum height=5mm,rounded corners=2pt,text depth=.25ex},
  long_rec/.style={rec, minimum width=4.3cm},
  long_rec_no_corner/.style={rec, minimum width=4cm,rounded corners=0,thin},
  arr/.style={-stealth, very thick},
  plus/.style={draw,circle,inner sep=0},
  ]

  \definecolor{mypara}{RGB}{197,193,230}
  \definecolor{myyellow}{RGB}{255,217,101}
  \definecolor{myblue}{RGB}{0,112,192}

  % bottom input
  \node[long_rec, fill=mypara] (input) {Surrounding Sentence Representation };

  % Multi-head context attention
  \node[long_rec,above=2ex of input,fill=myyellow] (mh) {Multi-Head Context Attention};

  \draw[arr] (input) to node[txt,right] {$S\in\Real^{d\times n}$} (mh);

  \coordinate[left=1.2 of mh.north east] (mh_r);
  \coordinate[right=1.2 of mh.north west] (mh_l);

  % local visible pooling
  \node[rec,above=2ex of mh_r,fill=myyellow] (lvp) {Local Visible Pooling};

  \draw[arr] (mh_r) to node[txt,right] {$H\in\Real^{d\times n}$} ( lvp);

  %% arrow to the r side
  \draw[->,densely dashed,thick] (lvp.east) -- ++(1cm,-5mm);

  % relative position encoding
  \node[plus,above=1.25 of mh_l] (plus_l) {$+$};
  \node[plus,above=1.25 of mh_r] (plus_r) {$+$};

  \node[rec,fill=myblue,text=white,txt] at ($(plus_l)!0.5!(plus_r)$) (rpe) {Relative\\Position\\Encoding};

  \draw[arr] (mh_l) -- (plus_l);
  \draw[arr] (lvp) -- (plus_r);
  \draw[arr] (rpe) -- (plus_l);
  \draw[arr] (rpe) -- (plus_r);

  % global update
  \node[long_rec_no_corner,above=1ex of rpe, fill=myyellow] (mhc)
  {Multi-Head Blank-Context Attention};

  \node[long_rec_no_corner,above=0 of mhc, fill=myyellow] (gsu)
  {Gated Sequential Update};

  \node[long_rec,fit={(mhc) (gsu)}, inner sep=1pt] (global) {};
  \node[left=0 of global] {$N\times$};

  \draw[arr] (plus_l) to node[txt,left] {$H\in\Real^{d\times n}$} (plus_l |- global.south);

  \draw[arr] (plus_r) to node[txt,right] {$\bb^{(0)}\in\Real^d$} (plus_r |- global.south);

  %% arrow to the r side
  \draw[->,densely dashed,thick] (global.east) to node[txt,above,sloped]
  {global\\update} ++(1cm,0);

  % top layer
  \node[long_rec,above=2.1ex of gsu, fill=mypara] (be) {Blank Embedding $\bb\in\Real^d$};

  \draw[arr] (global) to node[txt,right] {$\bb^{(N)}\in\Real^d$} (be);

\end{tikzpicture}

%%% Local Variables:
%%% mode: latex
%%% TeX-master: "main"
%%% End:
}
  \subfloat[Local visible pooling and global update\label{fig:local_global}]{%
    \includegraphics[width=.56\textwidth]{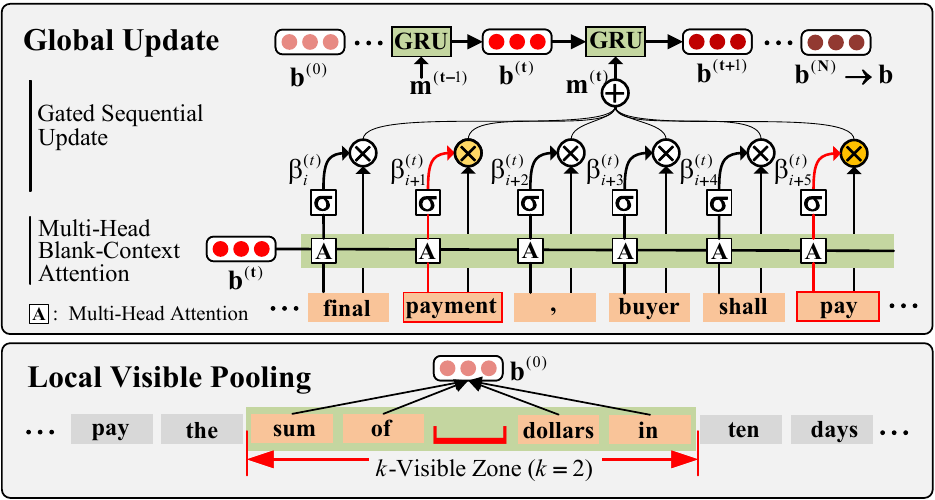}}
  \caption{The architecture of \texttt{BlankCoder}.
  It extends the Transformer encoder architecture to model meaningless blanks.
  First, we perform local visible pooling to build an initial blank embedding $\bb^{(0)}$ with the more related local context information. Second, we
  perform global update to recurrently refine the initial blank embedding by gathering and injecting relevant descriptions from the global context words.}
  \label{fig:model}
\end{figure*}

Figure~\ref{fig:model} depicts the architecture of \texttt{BlankCoder}, which mainly
consists of three modules: 1) {\em multi-head context attention}, which
associate relevant context words, 2) {\em local visible pooling}, which
initiates the blank representation with local keywords, and 3) {\em global update}, which refine and update the blank representation with related context
words.

In the following, we focus on describing how to obtain the embedding of a blank
$b\in B$.

\subsubsection{Surrounding Sentence Representation}

Given a blank $b\in B$ and its surrounding sentence $s$ consisting of $n$ words
and one blank (i.e., $|s|=n+1$), we embed the $i$-th word $w_i\in s$ to a
$d$-dimension vector $\be_i\in\Real^d$.
Embeddings $\{\be_1,\ldots,\be_n\}$ are stacked up to form a matrix
$S\in\Real^{d\times n}$, which will be the representation of the surrounding
sentence.
Embedding vector $\be_i$ is the sum of three different embedding vectors capturing
different semantic aspects.

The first embedding vector is the {\em word embedding}, denoted by
$\be_i^\mathrm{word}\in\Real^d$, which is obtained by using a pre-trained word2vec
model.
Word embedding $\be_i^\mathrm{word}$ only contains the linguistic information of
word $w_i$.

The second embedding vector is the {\em positional embedding}, denoted by
$\be_i^\mathrm{pos}\in\Real^d$, which incorporates the order information of word
$w_i$ in surrounding sentence $s$.
We calculate $\be_i^\mathrm{pos}$ using the same approach in the vanilla
Transformer model~\cite{Vaswani2017}.

The third embedding vector is the {\em segmentation embedding}, denoted by
$\be_i^\mathrm{seg}\in\Real^d$, which distinguishes the preceding and following context words of the blank.
This can be viewed that the blank separates the surrounding sentence into two
segments, and each segment has one segmentation embedding.
We adopted the learned segmentation embeddings as implemented in BERT~\cite{Devlin2019}.

Following BERT, the final embedding of the $i$-th word $w_i\in s$ is calculated by
$\be_i\triangleq \be_i^\mathrm{word} + \be_i^\mathrm{pos} + \be_i^\mathrm{seg}$.

\subsubsection{Multi-Head Context Attention}
To adequately incorporate relevant descriptions for blank modeling, it’s essential to first model contextual information and correlate relevant words in the surrounding sentence.
To this end, in the \texttt{BlankCoder}, we first adopt the powerful multi-head self-attention mechanism as implemented in the Transformer encoder for contextual encoding.

The Transformer encoder takes the embedding matrix $S$ as input, and updates the
embedding matrix layer by layer.
The $i$-th layer outputs a matrix $H^{(i)}\in\Real^{d\times n}$ by
\[
  H^{(i)} = \texttt{FFN}(\texttt{MultiHeadAtt}(H^{(i-1)})),\quad  H^{(0)}=S.
\]
Here, \texttt{FFN} represents a feed-forward neural network, and
\texttt{MultiHeadAtt} is the multi-head version self-attention, i.e.,
\begin{align*}
  & \texttt{SelfAtt}(S) = V\cdot\mathrm{softmax}(\frac{K\transpose Q}{\sqrt{d_k}}) \\
  & Q=W_qS,\quad K=W_kS,\quad V=W_vS.
\end{align*}
Here $Q\in\Real^{d_k\times n},K\in\Real^{d_k\times n}$, and $V\in\Real^{d\times
  n}$ denote the query, key and value matrix, respectively.
$W_q\in\Real^{d_k \times d}$, $W_k\in\Real^{d_k \times d}$, and $W_v \in \Real^{d
  \times d}$ are learnable parameters.
We refer to~\cite{Vaswani2017} for more details on the Transformer encoder.

The output of the last Transformer encoder layer, denoted by $H\in\Real^{d\times
  n}$, is a better representation of the surrounding sentence, that incorporates
the linguistic information, position information, and context information of the
surrounding sentence.
In the following, we discuss how to obtain the blank embedding using the surrounding
sentence embedding $H$.

\subsubsection{Local Visible Pooling}

Because a blank belongs to its surrounding sentence, a straightforward way to
obtain the blank embedding is to sum the embeddings of all the words in the
surrounding sentence, i.e., sum the columns of $H$.
However, words in the surrounding sentence may have different importance for
describing the blank, and simply summing them ignores their difference and may
include too much noise.

Intuitively, the words that are close to the blank should be more related to the
blank than words that are faraway in the surrounding sentence.
Similar observation has been reported in other language modeling
tasks~\cite{Yin2016,Aloraini2020}.
Therefore, we propose to sum embeddings of these closer words to obtain an {\em
  initial blank embedding}, which will be further refined.
We refer to this step as {\em local visible pooling}, as illustrated in
Figure~\ref{fig:model}.

Ideally, we want to find a vector $\balpha\in\Real^n$ such that it has larger
values for words in a small {\em local region} around blank $b$ and almost zero
values for words outside of this region.
Then, blank $b$'s {\em initial embedding} $\bb^{(0)}\in\Real^d$ can be represented
as
\[
  \bb^{(0)} = H\balpha.
\]
We refer to $\balpha$ as the {\em pooling weights}, and we discuss how to obtain
pooling weights $\balpha$ In the following.

We first formally define this local region.
Assume blank $b$ is between the $i$-th and $(i+1)$-th word in the surrounding
sentence, $0\leq i\leq n$.\footnote{Here $i=0$ (or $i=n$) means that the blank is
  at the beginning (or end) of the surrounding sentence.}
We define the {\em $k$-visible zone} of blank $b$ as the set of (at most) $2k$
closest positions around $b$, i.e.,
\[
  \mathrm{zone}_k\triangleq \{\max(1, i-k+1), \ldots, \min(n,i+k)\}.
\]
In our design, only words in this $k$-visible zone have non-zero pooling weights,
and even words in the $k$-visible zone may have different pooling weights,
depending on how informative the word could describe the blank.
We propose a {\em masked self-attention} to calculate $\balpha$, i.e.,
\[
  \balpha = \mathrm{softmax}(\tanh(WH)\transpose\bq_w + \bu).
\]
Here $\bu\in\Real^n$ is a mask vector and $\bu_i=0$ if $i\in\mathrm{zone}_k$;
$\bu_i=-\infty$ otherwise.
$\bq_w\in\Real^{d_w}$ and $W\in\Real^{d_w\times d}$ will be learned during
training.
That is, surrounding sentence representation $H$ is first fed to a one-layer
perceptron to obtain its hidden representation.
Then we measure the similarity between the hidden representation and vector
$\bq_w$.
$\bq_w$ can be viewed as a fixed query that queries important words over a set of
words represented by the hidden representation.
Therefore, our design will assign those words that are close and informative to
the blank larger pooling weights than faraway and meaningless words in the
surrounding sentence.

\noindent\textbullet\,\textbf{Remark}.
The blank $b$ now has an initial embedding $\bb^{(0)}$, and words already have
embedding matrix $H$.
At this time, we suggest adding the {\em relative position embeddings} (RPEs) to
these embeddings to incorporate the relative position information between words
and the blank.
RPEs are directly adapted from the PEs in Transformer: the blank's RPE is simply
$\mathbf{PE}_0$; if two words have the same distance $x$ to the blank, then the
two words have the same RPE, i.e., $\mathbf{PE}_x$.
We will slightly abuse the notations and still use $\bb^{(0)}$ and $H$ to denote
the embeddings that have been added RPEs.

\subsubsection{Global Update}

The local visible pooling step only uses blank $b$'s $k$-visible zone in the
surrounding sentence to obtain an initial embedding $\bb^{(0)}$.
Note that some words outside of the $k$-visible zone may be also informative and
useful for describing the blank.
We propose a {\em global update} step that uses words in the entire surrounding
sentence to further refine the initial embedding $\bb^{(0)}$.
After global update, blank embedding will incorporate more contextual information
in the surrounding sentence, and hence will be more representative than
$\bb^{(0)}$.

The global update step refines initial embedding $\bb^{(0)}$ in a recurrent
manner, and we denote the blank embedding after $t$-th refinement by $\bb^{(t)}$.
In the following, we describe how to refine $\bb^{(t)}$ and obtain $\bb^{(t+1)}$, at
the $t$-th global update step.

Given the current blank representation $\bb^{(t)}$ and surrounding sentence
embedding $H$, we want to find words in the surrounding sentence that are
correlated to the blank.
This can be achieved by computing a correlation score between the blank and a
word, and we propose to compute the correlation score in the following way, i.e.,
\[
  \bbeta^{(t)}=\sigma\left(\frac{K\bq^{(t)}}{\sqrt{d_k}}\right)\,\,
  \text{where}\,\, \bq^{(t)} = W_{uq} \bb^{(t)}, K = W_{uk}H
\]
Here $\sigma$ denotes a sigmoid function, $W_{uq}\in\Real^{d_k\times d}$ maps vector
$\bb^{(t)}$ to a query vector, $W_{uk}\in\Real^{d_k\times d}$ maps matrix $H$ to a
key matrix, and $\bbeta^{(t)}\in\Real^n$ is a scaled inner-product vector
measuring the correlation score between the blank and each word in the surrounding
sentence.
In practice, we use a multi-head mechanism to calculate $\bbeta^{(t)}$ which will
capture similarities from different aspects, similar to Transformer.
\[
  \bbeta^{(t)} = \texttt{MultiHeadScore}(\bb^{(t)}, H)
\]
Note that we do not normalize elements of $\bbeta^{(t)}$ jointly by a softmax
operation, as is used in Transformer.
Instead, we normalize elements of $\bbeta^{(t)}$ individually by a sigmoid
operation because it is possible that none of the words in the surrounding
sentence is informative for describing the blank.
We refer to the above processing by {\em multi-head blank-context attention}.

Score vector $\bbeta^{(t)}$ will be used as a memory gate to guard whether we
should memorize the corresponding vector in $H$ or not, i.e.,
\[
  \bbm^{(t)}  = H\bbeta^{(t)}
\]
where $\bbm^{(t)}\in\Real^d$ is the memorized embedding vectors in $H$ and those
memorized vectors are correlated to the blank.
Hence $\bbm^{(t)}$ is considered to be informative for describing the blank and
should be used to refine $\bb^{(t)}$.
To this end, we propose to apply the classic GRU~\cite{Cho2014a} which takes
$\bbm^{(t)}$ as input and $\bb^{(t)}$ as hidden state, i.e.,
\[
  \bb^{(t+1)} = \texttt{GRU}(\bbm^{(t)}, \bb^{(t)}).
\]

The above procedure will repeat $N$ times.
Thus the initial blank embedding $\bb^{(0)}$ will be refined recurrently.
We refer to the processing as the {\em gated sequential update}.
The final output $\bb^{(N)}$ will be considered as the final embedding of blank
$b$, i.e., $\bb\triangleq\bb^{(N)}$.
The multi-head blank-context attention operation and gated sequential update
operation are illustrated in Figure~\ref{fig:model}.

\subsection{Classifier}

Armed with the elaborately designed \texttt{BlankCoder}, we are now able to
predict the consistency relation between two blanks following the routing in
Figure~\ref{fig:PBR}.

Given two blanks $b_i$ and $b_j$, we first obtain their embeddings $\bb_i$ and
$\bb_j$ using the \texttt{BlankCoder}, respectively.
Then we concatenate $\bb_i$, $\bb_j$, and also a comparative term $|\bb_i-\bb_j|$
as the representation of the input blank-pair.
We adopt a feed-forward neural network-based binary classifier to predict the
consistency relation $\hat{r}_{i,j}$, i.e.,
\[
  \hat{r}_{i,j} = \texttt{FFN}([\bb_i : \bb_j : |\bb_i - \bb_j|]) \in [0,1].
\]
Here $:$ denotes vector concatenation, $|\cdot|$ denotes element-wise difference,
and \texttt{FFN} is a feed-forward neural network that the middle layer activation
functions are ReLUs and the last layer activation function is sigmoid.
The comparative term works as an implicit contrastive loss that forces consistent
blanks to have similar representations and vice versa.

During training, we choose the Focal Loss~\cite{Lin2017Focal} as the objective to
address the class imbalance issue in the datasets.
% Focal Loss is a variant of the standard cross-entropy loss that down-weights the
% loss assigned to well-classified examples.
\[
  \mathrm{FL}(\hat{r}_{i,j}, r_{i,j}) =
  \begin{cases}
    - \alpha(1 - \hat{r}_{i,j})^\gamma \log(\hat{r}_{i,j}), & r_{i,j}=1,\\
    - (1-\alpha)\hat{r}_{i,j}^\gamma \log(1 - \hat{r}_{i,j}), & r_{i,j}=0.
  \end{cases}
\]
Here, $r_{i,j}\in\{0,1\}$ is the ground-truth consistency relation, $\alpha$ and
$\gamma$ are the balanced factor and focusing parameter of the focal loss,
respectively.

\section{Experiments}

In this section, we conduct experiments on two real-world contract datasets to
evaluate our method.

\subsection{Contract Datasets}

\bullethdr{Chinese Contracts}
We have collected $246$ open source business contract templates and $35$ real
contracts from a company.
These contracts are written in Chinese, and cover categories such as investment,
lease, and labor.
Annotation details of templates are reported in our technical appendix.
% For contract templates, we recruited three law students to fill them with proper
% content.

\bullethdr{ICAIL Contracts} by Chalkidis et al.~\shortcite{Chalkidis2017}.
These contracts are written in English, and they were used for the task of tagging
contract elements.
We treat these extracted elements as filled blanks.
The contracts are anonymized where the words
% including punctuations 
are replaced with numbers for privacy concerns.

For each contract document, we collect all the blank pairs, and obtain their
consistency relation by comparing the filled content.
This allows us to build large training datasets.
However, the class labels are imbalanced. For the Chinese Contracts, the
ratio between positive samples and negative samples is $1:59$, and for the ICAIL
Contracts, the ratio is $1:48$.
The statistics of these datasets are summarized in Table~\ref{tab:data}.

\begin{table}[htp]
\centering
\small
\begin{tabular}{r|rrr}
\hline
\hline
  contract dataset  & \#~contracts & \#~blank-pairs & pos : neg \\
  \hline
  Chinese Contracts & $281$       & $299,621$     & $1:59$    \\
  \hline
  ICAIL Contracts   & $1,526$     & $67,765$      & $1:48$    \\
  \hline
\end{tabular}
\caption{Data statistics}
\label{tab:data}
\end{table}

\begin{table*}[htp]
\small
\centering
\begin{tabular}{@{\ \ }l@{\ \ }|@{\ \ }c@{\ \ }c@{\ \ }c@{\ \ }c@{\ \ }c@{\ \ }c|c@{\ \ }c@{\ \ }c@{\ \ }c@{\ \ }c@{\ \ }c@{\ \ }}
\hline
\hline
                     & \multicolumn{6}{c|}{Chinese Contracts} & \multicolumn{6}{c}{ICAIL Contracts}  \\
\hline
   model & AUC    & accuracy    & precision    & recall    & F1    & MCC    & AUC    & accuracy    & precision    & recall    & F1    & MCC   \\
\hline
BiLSTM               & $98.26$ & $92.58$ & $88.24$ & $86.11$ & $87.16$ & $86.96$   & $96.06$ & $84.86$ & $79.22$ & $70.11$ & $74.39$ & $74.03$ \\
Transformer          & $94.45$ & $88.51$ & $79.47$ & $77.36$ & $78.40$ & $78.04$   & $93.37$ & $83.09$ & $74.36$ & $66.67$ & $70.30$ & $69.82$ \\
Transformer-seg      & $97.15$ & $92.32$ & $85.15$ & $86.90$ & $86.01$ & $85.78$   & $95.26$ & $83.99$ & $77.78$ & $68.39$ & $72.78$ & $72.40$ \\
CenteredLSTM         & $98.35$ & $92.68$ & $88.96$ & $86.59$ & $87.76$ & $87.56$   & $96.15$ & $85.71$ & $78.62$ & $71.84$ & $75.08$ & $74.66$ \\
AttnLSTM             & $98.62$ & $92.85$ & $90.83$ & $85.51$ & $88.09$ & $87.93$   & $96.19$ & $85.76$ & $80.92$ & $70.69$ & $75.46$ & $75.16$ \\
CorefBERT            & $93.06$ & $90.20$ & $53.85$ & $81.58$ & $64.87$ & $65.61$   & $-$ & $-$ & $-$ & $-$ & $-$ & $-$ \\
\hline
{\bf PBR}       & $\textbf{98.73}$ & $\textbf{94.05}$ & $\textbf{93.74}$ & $\textbf{88.22}$ & $\textbf{90.90}$ & $\textbf{90.77}$   	& $\textbf{96.25}$ & $\textbf{86.01}$ & $\textbf{81.22}$ & $\textbf{72.08}$ & $\textbf{76.38}$ & $\textbf{76.09}$ \\
{\%Improvement}       & $0.11$ & $1.29$ & $3.20$ & $1.88$ & $3.19$ & $3.23$   	& $0.06$ & $0.29$ & $0.37$ & $0.33$ & $1.22$ & $1.24$ \\
\hline
\end{tabular}
\caption{Model evaluation results (\%).
  Note that the accuracy above stands for the balanced accuracy.
  \label{tab:overall}}
\end{table*}

\subsection{Settings}

\subsubsection{Evaluation Metrics.}
We use AUC, precision, recall, F1 score, {\em balanced accuracy}, and {\em Matthews
correlation coefficient} (MCC)~\cite{Boughorbel2017} as evaluation metrics.
Balanced accuracy is defined as the average of recall obtained in each class.
MCC is a correlation coefficient between the predicted and ground truth
binary classifications, and it has a value between $-1$ and $+1$, with $+1$
representing a perfect prediction.
Note that because the datasets are highly imbalanced, we thus focus on comparing balanced measures: F1 and MCC.

\subsubsection{Baselines}
We compare our method with the following approaches in text infilling and zero pronoun
resolution.
\begin{itemize}
\item \textbf{BiLSTM}~\cite{graves2013speech} is a widely adopted bidirectional
  RNN in text infilling~\cite{fedus2018maskgan,liu2019tigs}, where a blank is
  treated as an out-of-vocabulary token in the sentence and modeled together with
  the context words.

\item \textbf{Transformer}~\cite{Vaswani2017} is a SOTA attention based language representation
  model.
  We adopt the vanilla Transformer encoder and perform
  blank modeling in the same way as in BILSTM.

\item \textbf{Transformer-seg}~\cite{zhu2019text} is a segment-aware Transformer
  for text infilling, where the segments are obtained by splitting the sentence
  at the blanks.

\item \textbf{CenteredLSTM}~\cite{yin2017chinese} is a zero pronoun modeling
  method that adopts two RNNs with one encoding the preceding context sequentially
  and the other encoding the following context reversely.
  A blank is represented with the concatenation of the two last hidden vectors.

\item \textbf{AttnLSTM}~\cite{yin2018zero} extends CenteredLSTM with a
  self-attention mechanism to generate segment representations that are
  concatenated to encode the blank.

\item \textbf{CorefBERT}~\cite{Aloraini2020} is a zero pronoun modeling method
  that adopts a pre-trained BERT \cite{Devlin2019} for contextual encoding and
  represents a blank by averaging the embeddings of its nearest two words.

\end{itemize}

\begin{table}[htp]
  \centering
  \small
  \begin{tabular}{@{\ \ }l@{\ \ }|@{\ \ }c@{\ \ }c@{\ \ }c@{\ \ }c@{\ \ }c@{\ \ }c@{\ \ }}
    \hline
    \hline
    Metrics    & AUC     & accuracy & precision & recall  & F1      & MCC     \\
    \hline
    PBR        & $98.73$ & $94.05$  & $93.74$   & $88.22$ & $90.90$ & $90.77$ \\
    -no local  & $97.76$ & $92.31$  & $81.51$   & $87.86$ & $84.57$ & $84.36$ \\
    -no update & $98.14$ & $92.95$  & $87.81$   & $86.11$ & $86.95$ & $86.74$ \\
    -no cmp    & $98.38$ & $93.43$  & $93.57$   & $86.96$ & $90.14$ & $90.04$ \\
    \hline
  \end{tabular}
  \caption{Ablation analysis on the Chinese Contracts.\label{tab:ablation}}
\end{table}

\subsubsection{Implementation Details}
We implement all the benchmarks using Pytorch on a server equipped with 2 Nvidia
Tesla V100 GPUs, each with 32GB memory.
For the Chinese Contracts, we adopt the Chinese word2vec embeddings released by Li
et al.~\shortcite{Li2018}.
For the ICAIL Contracts, the word2vec embeddings are provided by Chalkidis et
al.~\shortcite{Chalkidis2017}.
We perform the hyper-parameter search to find the best combinations for all the
models.
The details are shown in the appendix.\footnote{Codes available at \url{https://github.com/ShuoZhangXJTU/CIC}}

\subsection{Model Evaluation}

We compare the performance of different models, and show the results in
Table~\ref{tab:overall}.
Note that CorefBERT is not applicable on the ICAIL Contracts due to data
anonymization.

We observe that our PBR framework outperforms the others in terms of all
evaluation metrics.
Specifically, for balanced accuracy, PBR achieves $0.29\sim 1.29\%$ improvements,
for F1 score, PBR achieves $1.22\sim 3.19\%$ improvements, and for MCC, PBR
achieves $1.24\sim 3.23\%$ improvements.

Compared to the vanilla Transformer encoder and its segment-aware version, our
\texttt{BlankCoder} in PBR also has significant improvement.
The balanced accuracy is improved by $1.87\sim 6.26\%$, the F1 score is improved
by $4.95\sim 15.94\%$, and MCC is improved by $5.09\sim 16.31\%$.
This demonstrates the effectiveness of the two-stage semantic modeling approach in
our \texttt{BlankCoder}.

We also observe a general performance decline of all the benchmarks and a limited
performance gain of our PBR framework on the ICAIL Contracts.
We attribute this to data anonymity and the ambiguous blank pair samples. 
In the ICAIL dataset, the sentences are not split to fit the task of tagging
contract elements. 
Due to its anonymity, we have to specify a region as a pseudo-sentence, which
would contain incomplete or redundant sentences that hinder the performance.
Besides, ambiguous blank pairs where both the same and different contents are allowed can
introduce extra noise that impairs the performance.
Details are illustrated in our technical appendix due to space limitations.
% We attribute this to its anonymity and the incompatible data format of the
% dataset.
% In the ICAIL dataset, the sentences are not split to fit the task of tagging
% contract elements.
% Due to its anonymity, we have to specify a region as a pseudo-sentence, which
% would contain incomplete or redundant sentences that hinder the performance.
% \zz{In addition, in real-world contracts, there can be blank pairs for which both same and different content are allowed
% (e.g., Name and corresponding initials).  
% Since the ICAIL data is labeled by comparing filled contents, the ambiguous blank pairs can introduce extra noise and impair performance of all the models, while such cases are avoided in the Chinese Contracts dataset through our labeling strategy (see Appendix).}

\begin{figure}[htp]
  % \centering
  \includegraphics[width=.42\textwidth]{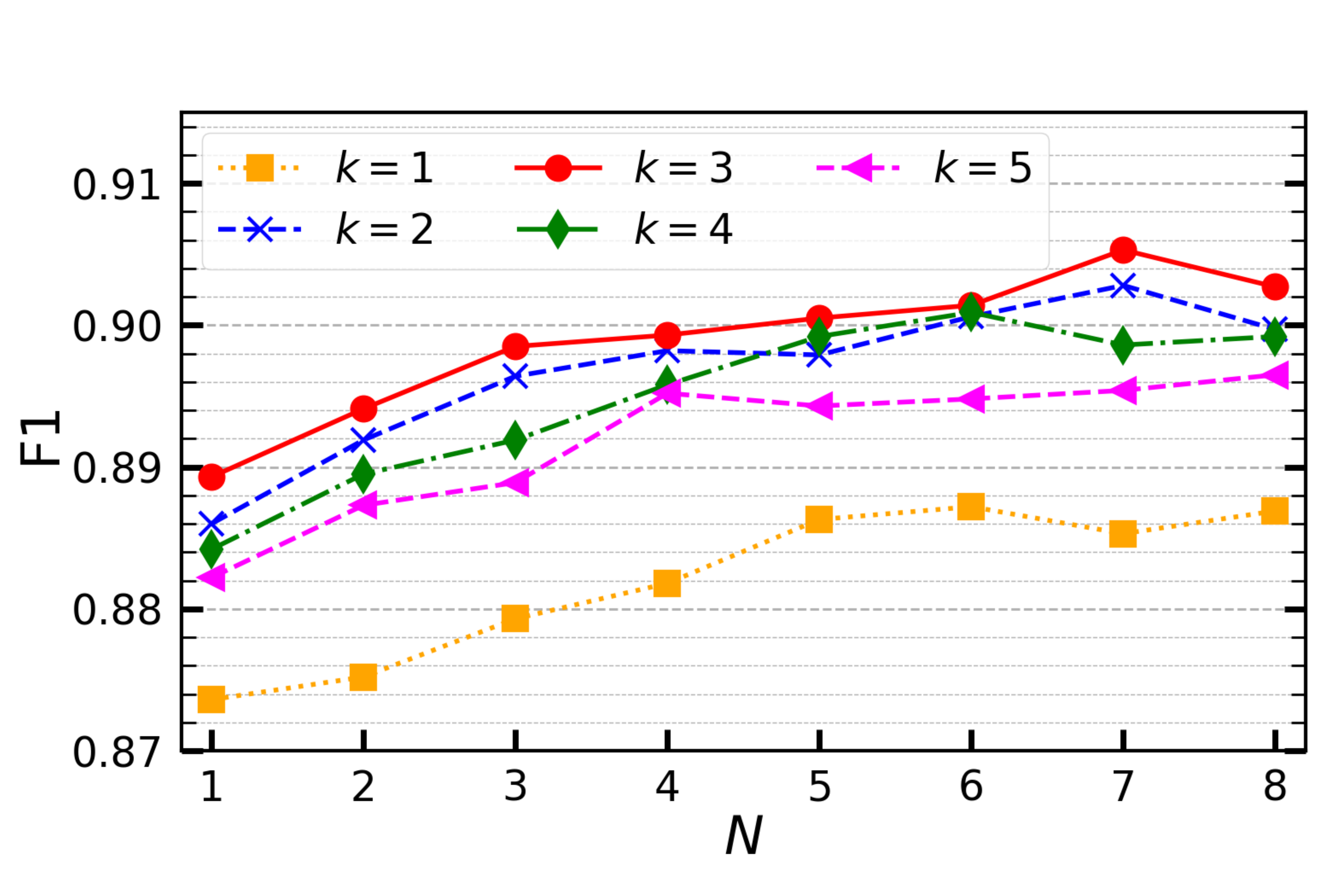}
  \caption{Hyper-parameter sensitivity \label{fig:comarative}}
\end{figure}

\subsection{Ablation Experiments}

To further investigate the performance of the two-stage blank modeling strategy,
we conducted ablation experiments on PBR (see Table~\ref{tab:ablation}).
Similar results on the ICAIL Contracts are omitted because of space limitations.

To verify the effectiveness of the local visible pooling layer, we build a
no-local version PBR (i.e., ``-no local''), which directly assigns the whole
sentence as the local visible zone.
To evaluate the importance of the word-level semantic reasoning, we remove global
update (i.e., ``-no update'').
To evaluate the effectiveness of the comparative term (i.e., $|\bb_i-\bb_j|$), we
build a base PBR with no comparative term in the classifier (i.e., ``-no cmp'').

In Table~\ref{tab:ablation}, we observe that both the local visible pooling and
global update are important components of the PBR framework.
% , and they can firmly affect the performance of PBR.
The local visible pooling is more critical since it determines if there exists
noise information in the initial embedding.
We also observe that the absence of the comparative term in the classifier leads
to performance decrease, which implies the importance of contrastive supervision.

\subsection{Hyper-parameter Sensitivity}

In this section, we study how hyper-parameters in BlankCoder affect the
performance, i.e., the length of local visible zone $k$, and the number of the
stacked global update blocks $N$.
The results on the Chinese Contracts are shown in Figure~\ref{fig:comarative}.
We omit similar results on the ICAIL Contracts due to space limitations. 

In Figure~\ref{fig:comarative}, we show five different lines where each line
denotes the performance (F1) using different $N$ and $k$.
Fixing $k$, we observe that the stack of multiple global update blocks could
consistently improve the performance, which demonstrates the effectiveness of the
recurrent update process.
Fixing $N$, we observe that the performance increases with the size of the local
region at first and then decreases.
This is due to the fact that a minimal local region does not contain enough
related information, while a large local region would introduce irrelevant noise
words.
To reach better performance, a proper local visible zone shall be determined for the
\texttt{BlankCoder}.

\begin{figure}[htp]
\centering
\includegraphics[width=.40\textwidth]{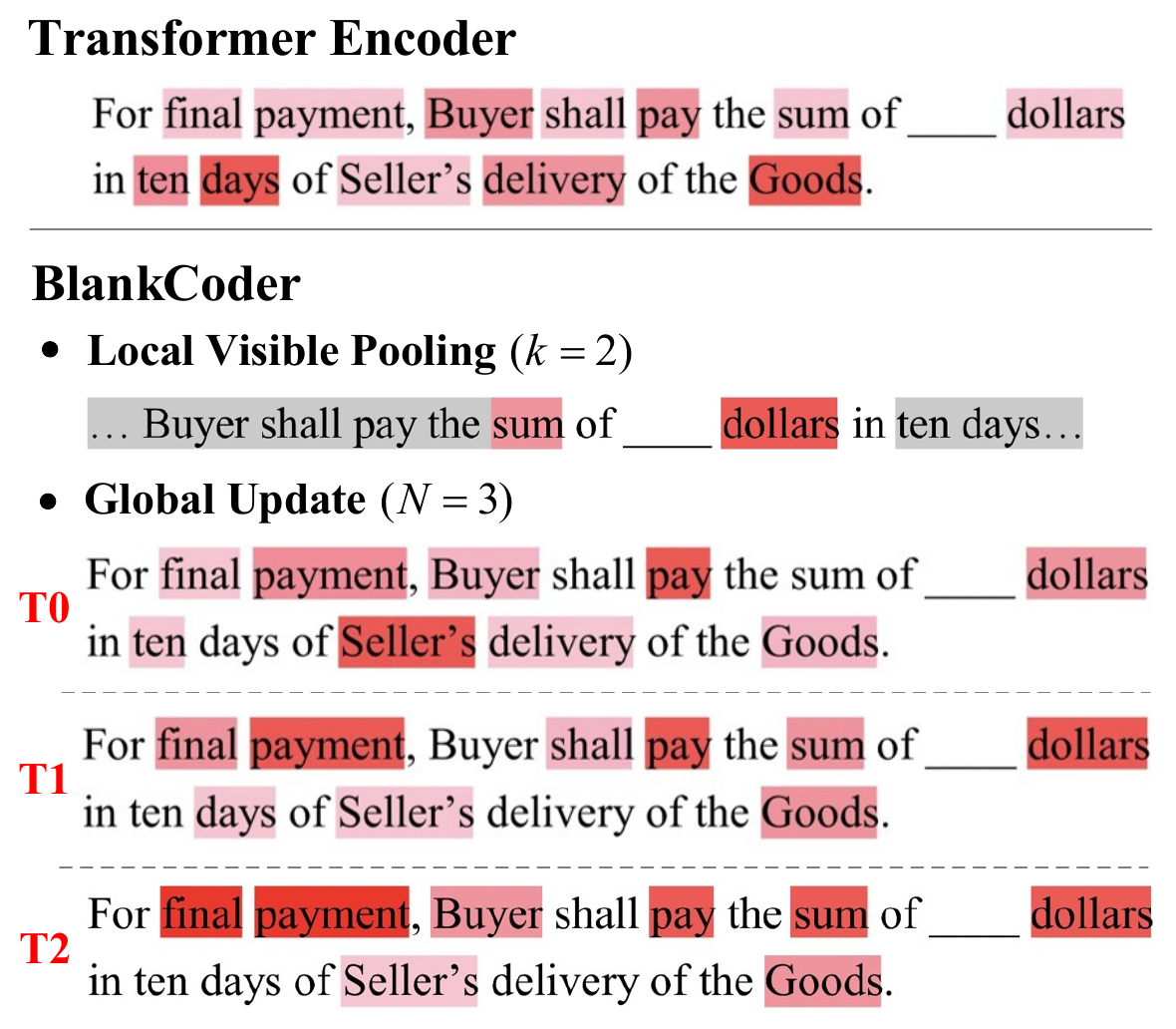}
\caption{The attention visualization on a case example in a real-word sales
  contract.}
\label{fig:case}
\end{figure}

\subsection{Case Study: Attention in \texttt{BlankCoder}}

To intuitively show the effectiveness that PBR can extract representative
features, we compare the attention of the Transformer encoder and our
\texttt{BlankCoder}.
For simplicity, we average the attention weights for all the Transformer encoder
layers and omit the context attention layers of our \texttt{BlankCoder}.

Figure~\ref{fig:case} shows a case example in a real-word sales contract with the
blank referring to the final payment amount, where darker red shades indicate
larger attention weights.
For the Transformer encoder, we see that irrelevant descriptions like ``days'' and
``delivery'' are incorrectly highlighted, which would lead to biased blank
representation that hinders further consistency checking.
Whereas in our \texttt{BlankCoder}, the blank is first initialized with related
local keywords (e.g., ``dollars'') and then recurrently updated with other
relevant descriptions (e.g., ``final'' and `` payment''), yielding an accurate and
expressive blank representation without noise information.

\section{Conclusion and Future Work}

In this work, we formulate the {\em Contract Inconsistency Checking} (CIC)
problem, an automatic contract analysis task with significant practical
importance, and we propose a novel end-to-end {\em Pair-wise Blank Resolution}
(PBR) framework to predict the consistency relation for every two blanks with high
accuracy.
In PBR, we extend the Transformer encoder architecture and propose
\texttt{BlankCoder}, an off-the-shelf effective blank modeling method that could
easily generalize to other tasks such as text infilling.
Extensive experiments show that our model can significantly and consistently
outperform existing baselines, yielding a promising balanced accuracy of $94.05\%$ and an F1
score of $90.90\%$.
In the future, we plan to consider more complex cases (e.g., ambiguous blank pairs) and explore more complex consistency checking scenarios that require logical reasoning.
%  (e.g., The total price should be the sum of the
% individual prices).

\section{Acknowledgments}
The research presented in this paper is supported in part by National Key R\&D Program of China (2018YFC0830500), National Natural Science Foundation of China (61922067, U1736205, 61902305), MoE-CMCC “Artifical Intelligence” Project (MCM20190701), Natural Science Basic Research Plan in Shaanxi Province of China (2019JM-159), Natural Science Basic Research Plan in Zhejiang Province of China (LGG18F020016).

% \clearpage
\bibliographystyle{aaai21}
\bibliography{ref.bib}

\end{document}